\documentclass[letterpaper]{article} 
\usepackage[preprint]{aaai2027}  
\usepackage[hyphens]{url}  
\usepackage{graphicx} 
\usepackage{natbib}  
\usepackage{caption} 
\usepackage{algorithm}
\usepackage{algorithmic}

\usepackage{newfloat}
\usepackage{listings}
\DeclareCaptionStyle{ruled}{labelfont=normalfont,labelsep=colon,strut=off} 
\floatstyle{ruled}
\newfloat{listing}{tb}{lst}{}
\floatname{listing}{Listing}

\usepackage{booktabs}

\title{MEGRAG: Multi-Granular Evidence Graphs for Answer-Aware Multi-Hop RAG}
\author{
    Weidong Bao,
    Yingying Sun,
    Jun Yang,
    Yilin Wang,
    Zili Wei,\\
    Yubin Bao,
    Fangling Leng,
    Minghe Yu,
    Tiancheng Zhang,
    Ge Yu
}
\affiliations{
    Northeastern University, Shenyang, China\\
    baoweidong293@gmail.com,
    \{sunyingying,yangjun,weizl2\}@mails.neu.edu.cn,\\
    wangyilin0409@gmail.com,
    \{baoyubin,lengfangling,yuge\}@cse.neu.edu.cn,\\
    \{yuminghe,tczhang\}@mail.neu.edu.cn
}

\begin{document}

\maketitle

\begin{abstract}
Multi-hop question answering is a fundamental challenge in
retrieval-augmented generation (RAG), because deriving an
answer requires integrating dispersed evidence. Iterative RAG
(iRAG) is widely used for this challenge, but existing methods
have two limitations.
First, most methods still support each reasoning step with
single-granularity evidence, making it difficult to balance
information density and contextual noise. Second, existing
methods often answer the original question only after aggregating
evidence retrieved across intermediate steps, so redundant evidence
and intermediate retrieval errors may accumulate and degrade the
final answer. To address these limitations, we propose \textbf{MEGRAG}, an
answer-aware framework that represents multi-hop reasoning as a
path-structured multi-granular evidence graph. Offline, MEGRAG links
passages to their sentences and extracted triples through a
cross-granularity index. Online, it retrieves passages for the current
query and selects aligned evidence, starting with compact triples and
adding sentence or passage context as needed. MEGRAG uses the resulting
intermediate answer and prior reasoning to decide whether the Initial
Query has been resolved. If not, it identifies the missing information
and formulates a focused next query; otherwise, it stops retrieval and
returns the answer. Extensive experiments demonstrate consistent gains over a
diverse set of RAG baselines.

\end{abstract}


\section{Introduction}

Retrieval-Augmented Generation (RAG) performs strongly on simple knowledge-seeking queries and single-hop question answering~\cite{lewis2020retrieval,lin2024ra,ram2023context}. Multi-hop question answering is harder because relevant evidence is often dispersed across multiple sources and must be progressively integrated through reasoning~\cite{fan2024survey,trivedi2023interleaving,mallen2023not}. Standard single-step retrieval ranks evidence only against the initial query, often recovering local clues while missing intermediate evidence whose relevance emerges only after partial reasoning~\cite{shao2023enhancing}. Iterative RAG (iRAG) addresses this limitation by interleaving retrieval with reasoning and query reformulation, progressively refining the information need and gathering evidence for subsequent hops~\cite{trivedi2023interleaving,asai2024self,yao2025seakr}.

\begin{figure}[t]
    \centering
    \includegraphics[width=\columnwidth]{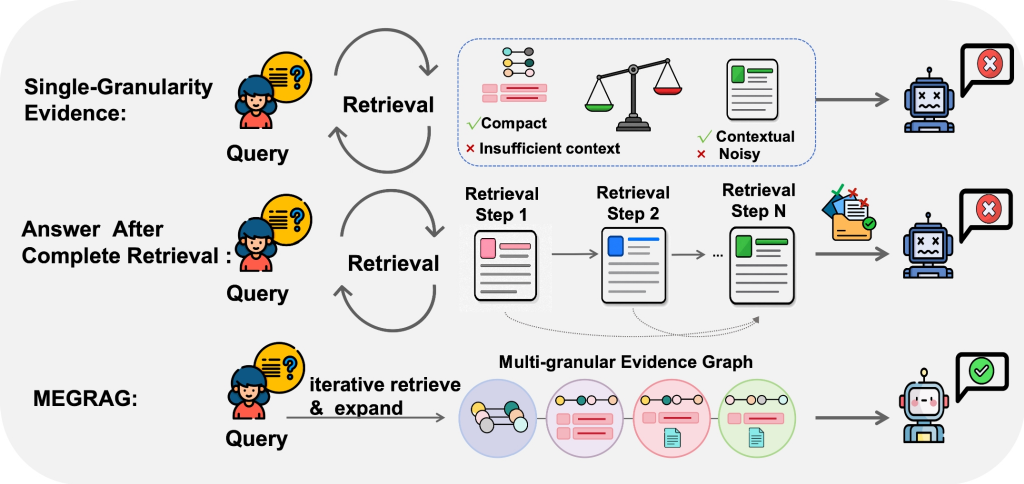}
    \caption{Motivation of MEGRAG. Fixed-granularity evidence can be
    insufficient or noisy, while answer-after-retrieval pipelines may
    accumulate redundant evidence and errors. MEGRAG builds multi-granular
    evidence at each step and uses its intermediate answer to continue or stop.}
    \label{fig:intro}
\end{figure}

Recent methods have advanced multi-hop RAG from three complementary perspectives. Iterative methods improve retrieval adaptivity by reformulating queries, diagnosing knowledge gaps, and progressively organizing retrieved information~\cite{zhou2024metacognitive,cheng2025dualrag}. Structure-aware methods strengthen connections among dispersed evidence and improve retrieval precision through graphs, hypergraphs, or evidence chains~\cite{gutierrez2025rag,wang2026cross,peng2026neocorrag}. Multi-granular methods better balance focused evidence with contextual completeness by using fine-grained units for passage ranking or adaptive context expansion~\cite{hu2026iterative,wei2026cirag}. Despite these advances, two limitations remain, as illustrated in Figure~\ref{fig:intro}. First, most methods still support each reasoning step with single-granularity evidence, making it difficult to balance information density and contextual noise: fine-grained evidence may lack sufficient context, whereas coarse-grained evidence may introduce irrelevant information. Second, most iterative methods answer the Initial Query only after completing retrieval and aggregating evidence from all intermediate steps. As a result, redundant evidence and intermediate retrieval errors may accumulate throughout the reasoning process and degrade the final answer.

Human reasoning is selective and goal-directed: people seek only enough
evidence to support a judgment, then use intermediate conclusions to identify
what remains unknown and guide
further inquiry~\cite{simon1955behavioral,nelson1990metamemory,loewenstein1994psychology}.
Motivated by this process, we propose MEGRAG, an answer-aware framework for
multi-hop RAG. Offline, MEGRAG builds a cross-granularity index that links each
passage to its sentences and extracted triples. Online, it retrieves passages
for the current query and selects aligned evidence, starting with compact
triples and adding sentence or passage context only as needed. MEGRAG uses the
selected evidence to answer the current query, then combines this intermediate
answer with prior reasoning to determine whether the Initial Query has been
resolved. If information is still missing, it identifies the remaining gap and
formulates a focused next query; otherwise, it stops retrieval and returns the
final answer. As this process unfolds, MEGRAG organizes the queries, selected
evidence, and intermediate answers into a question-specific, path-structured
evidence graph, with explicit transitions that record how each intermediate
answer reveals the next information need. Finally, the policy for constructing
this graph is distilled into a lightweight student.

The main contributions of this paper are summarized as follows:
\begin{itemize}

\item \textbf{A Flexible Multi-granular Evidence Framework.}
We propose MEGRAG, which separates reusable offline evidence organization from
question-specific online reasoning. Rather than representing the corpus as a
knowledge graph, MEGRAG organizes passage-, sentence-, and triple-level views
through a cross-granularity index and constructs a question-specific,
path-structured evidence graph online.

\item \textbf{Sufficiency-guided Multi-granular Evidence Selection.}
At each reasoning step, MEGRAG begins with compact triples and expands to aligned
sentences and passages only as needed, stopping at the first granularity judged
sufficient for the current query. This produces compact evidence without
assuming a fixed granularity across questions or reasoning steps.

\item \textbf{Answer-aware Iterative Reasoning.}
MEGRAG distinguishes answering the current query from resolving the Initial
Query. It identifies what remains missing and either formulates a focused next
query or stops retrieval. This iterative policy is distilled into a lightweight
student.

\end{itemize}

\section{Related Work}

\paragraph{Iterative and Structured Retrieval for Multi-hop Reasoning.}
Standard RAG retrieves once from the initial query, whereas iterative methods
make retrieval responsive to evolving information needs~\cite{lewis2020retrieval,trivedi2023interleaving}.
MetaRAG evaluates tentative answers to plan targeted refinement~\cite{zhou2024metacognitive},
and DualRAG couples reasoning-augmented querying with progressive knowledge
aggregation~\cite{cheng2025dualrag}. Structure-aware methods instead improve
evidence connectivity: HippoRAG~2 and HGRAG propagate relevance over corpus
structures~\cite{gutierrez2025rag,wang2026cross}, while LogicRAG, NeocorRAG,
and QAFD-RAG construct query-centered structures or chains
online~\cite{chen2026you,peng2026neocorrag,zhou2026query}. MEGRAG organizes
selected evidence and intermediate answers into an online path state that
explicitly records the remaining information need.

\paragraph{Multi-granular Evidence for RAG.}
Evidence granularity trades information density for contextual completeness:
passages preserve context but may contain noise, sentences retain local
constraints, and triples provide compact facts but may omit qualifiers. MGranRAG
uses sentence- and phrase-level evidence to recalibrate passage
ranking~\cite{hu2026iterative}. CIRAG iteratively integrates triple-centric
evidence and applies cascaded granularity to the accumulated context for final
answer generation~\cite{wei2026cirag}. MEGRAG instead constructs a
multi-granular evidence composition within each retrieval step, immediately
produces \(b_i\), and uses the resulting node state to determine
\(q_{i+1}\) or stop.

\begin{figure*}[t]
    \centering
    \includegraphics[width=\textwidth]{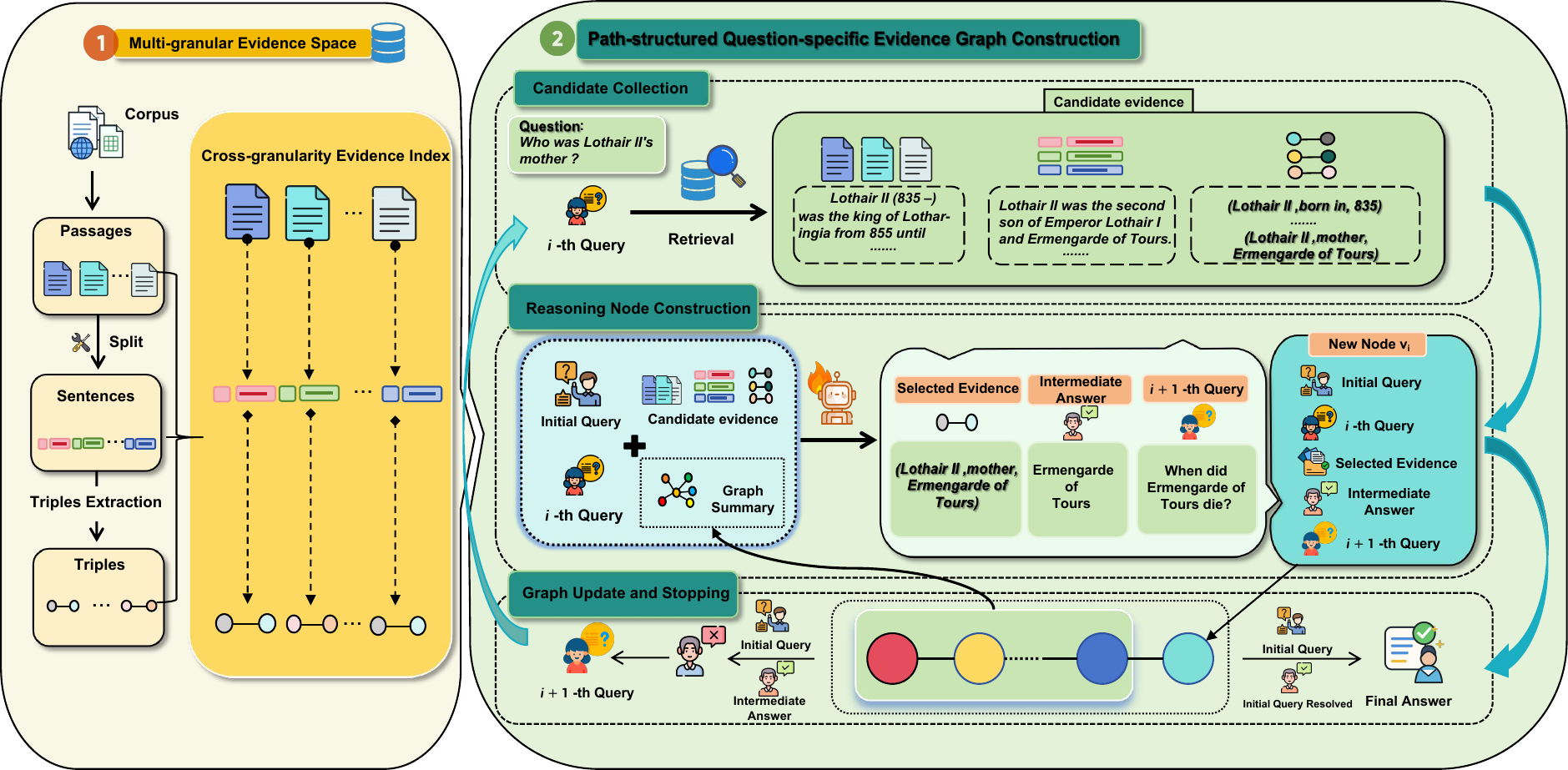}
    \caption{
    Overview of MEGRAG. The offline stage constructs aligned passage, sentence, and triple views. At step \(i\), the online system retrieves aligned candidates for \(q_i\), starts from compact triples, and adds sentences or passages when more context is needed to form \(Z_i\). It then derives an intermediate answer \(b_i\) and decides whether it can produce a terminal candidate \(y_i\) for the Initial Query. Otherwise it generates \(q_{i+1}\). The example distinguishes the Initial Query about a death date from the current query about the mother's identity.
    }
    \label{fig:framework}
\end{figure*}

\begin{table*}[t]
\centering
\small
\setlength{\tabcolsep}{1mm}
\renewcommand{\arraystretch}{1.10}
\begin{tabular*}{\textwidth}{@{\extracolsep{\fill}}lcccccccccccc@{}}
\toprule
& \multicolumn{6}{c}{\textbf{Qwen3-8B}}
& \multicolumn{6}{c}{\textbf{Qwen3-Max}} \\
\cmidrule(lr){2-7}\cmidrule(lr){8-13}
& \multicolumn{2}{c}{2WikiMQA}
& \multicolumn{2}{c}{HotpotQA}
& \multicolumn{2}{c}{MuSiQue}
& \multicolumn{2}{c}{2WikiMQA}
& \multicolumn{2}{c}{HotpotQA}
& \multicolumn{2}{c}{MuSiQue} \\
\cmidrule(lr){2-3}\cmidrule(lr){4-5}\cmidrule(lr){6-7}
\cmidrule(lr){8-9}\cmidrule(lr){10-11}\cmidrule(lr){12-13}
\textbf{Method}
& \textbf{F1} & \textbf{EM}
& \textbf{F1} & \textbf{EM}
& \textbf{F1} & \textbf{EM}
& \textbf{F1} & \textbf{EM}
& \textbf{F1} & \textbf{EM}
& \textbf{F1} & \textbf{EM} \\
\midrule
Direct Model
& 28.51 & 25.00 & 25.29 & 17.00 & 8.35 & 2.30
& 40.17 & 33.80 & 44.41 & 32.60 & 20.13 & 9.20 \\
NativeRAG
& 32.10 & 28.90 & 52.20 & 35.20 & 17.10 & 10.20
& 48.30 & 40.10 & 67.50 & 53.20 & 28.30 & 18.90 \\
\addlinespace[1.5pt]
DualRAG
& 63.10 & 51.90 & 59.20 & 44.90 & 34.30 & 26.19
& 76.20 & 66.83 & 74.14 & 58.80 & 51.60 & 37.50 \\
MetaRAG
& 52.30 & 45.80 & 64.33 & 50.45 & 32.90 & 23.60
& 59.10 & 53.40 & 75.80 & 61.40 & 44.80 & 33.60 \\
HippoRAG~2
& 66.94 & 56.83 & 64.83 & 49.68 & 38.64 & 27.39
& 74.18 & 64.72 & 72.09 & 58.64 & 53.68 & 41.76 \\
NeocorRAG
& 69.08 & 58.24 & 67.36 & 52.21 & 41.05 & 29.84
& 75.84 & 66.21 & 75.26 & 61.47 & 54.81 & 43.92 \\
CIRAG
& \underline{70.50} & \underline{61.20} & 68.10 & 53.50 & 42.10 & 30.10
& 77.30 & 68.10 & 75.20 & 61.10 & \underline{56.80} & \underline{45.50} \\
MGranRAG
& 60.73 & 52.60 & 73.79 & 57.80 & \underline{46.81} & 34.20
& 77.10 & 69.80 & 76.10 & \underline{63.50} & 52.20 & 40.60 \\
HGRAG
& 68.00 & 60.56 & \underline{74.05} & \underline{59.20} & 46.08 & \underline{35.80}
& \underline{78.30} & \underline{70.30} & \underline{76.40} & 63.30 & 53.80 & 42.20 \\
LogicRAG
& 70.21 & 59.62 & 65.74 & 50.31 & 40.36 & 29.18
& 77.08 & 67.86 & 72.81 & 59.35 & 55.47 & 44.06 \\
QAFD-RAG
& 66.34 & 59.00 & 73.60 & 59.00 & 43.96 & 33.90
& 66.11 & 59.10 & 76.28 & 60.90 & 48.32 & 38.20 \\
\midrule
\textbf{MEGRAG}
& \textbf{76.34} & \textbf{68.70} & \textbf{76.95} & \textbf{61.90} & \textbf{55.15} & \textbf{44.90}
& \textbf{80.22} & \textbf{71.40} & \textbf{79.74} & \textbf{64.70} & \textbf{62.01} & \textbf{51.80} \\
\bottomrule
\end{tabular*}
\caption{Main results on the same fixed 1,000-question subset of each multi-hop
QA benchmark. Baselines retain their originally specified training regimes:
trainable methods are trained as prescribed, while training-free methods remain
training-free. The Qwen3-8B block is therefore a complete-system comparison,
with MEGRAG including policy distillation; MEGRAG with Qwen3-Max is prompt-only.
The best result in each column is bold, and the strongest baseline is
underlined.}
\label{tab:main_multihop}
\end{table*}

\section{Methodology}

\subsection{Problem Formulation}

Given an Initial Query \(x\) and corpus
\(\mathcal{D}=\{d_j\}_{j=1}^{N}\), MEGRAG builds an online, question-specific
evidence graph \(G\) and initializes the current query as \(q_1=x\). After step
\(i\), the graph is
\begin{equation}
G^{(i)} = (V^{(i)}, E^{(i)}),
\end{equation}
where \(V^{(i)}\) contains reasoning nodes and \(E^{(i)}\) records transitions
between successive retrieval queries together with terminal stop edges. The
\(i\)-th node is
\begin{equation}
v_i=(q_i,Z_i,b_i,h_i,m_i,y_i,s_i).
\end{equation}
Here, \(Z_i\) is selected evidence, \(b_i\) is the evidence-grounded
intermediate answer to the current query \(q_i\), \(h_i\) is the resolved
reasoning path, and \(m_i\) is the remaining information need. The variable
\(y_i\) is the answer to the Initial Query \(x\) when \(s_i=1\), and
\(s_i\in\{0,1\}\) is the stop decision. When \(s_i=0\), \(y_i=\emptyset\),
\(m_i\neq\mathrm{none}\), and the policy generates \(q_{i+1}\) for \(m_i\).
For a continuing node (\(s_i=0\)), the corresponding transition edge is
\begin{equation}
e_i^{\mathrm{cont}}=(v_i,r_i,v_{i+1}),
\end{equation}
where \(r_i\) describes how the next query extends the current query toward
answering the Initial Query. Specifically, \(r_i\) is defined jointly from the
Initial Query \(x\), current query \(q_i\), and next query \(q_{i+1}\). A valid
stop has \(s_i=1\),
\(m_i=\mathrm{none}\), \(y_i\neq\emptyset\), and
\(q_{i+1}=\emptyset\). No \(v_{i+1}\) is constructed; instead, MEGRAG adds the
terminal edge
\begin{equation}
e_i^{\mathrm{stop}}=(v_i,\mathrm{STOP}).
\end{equation}

\subsection{Framework Overview}

Figure~\ref{fig:framework} summarizes the offline and online stages of MEGRAG.
Offline, passages are organized together with their sentence and triple views.
Online, MEGRAG retrieves passages for the current query, selects evidence at the
first granularity judged sufficient, and produces an intermediate answer. It
continues with a focused query when information is missing and returns the
answer once the Initial Query is resolved. The iterative policy is distilled
into a lightweight student.

\subsection{Multi-granular Evidence Space}

MEGRAG first constructs an offline multi-granular evidence space:
\begin{equation}
\mathcal{E}=\{\mathcal{E}^{P},\mathcal{E}^{S},\mathcal{E}^{T},\mathcal{I}\},
\end{equation}
where \(\mathcal{E}^{P}\), \(\mathcal{E}^{S}\), and \(\mathcal{E}^{T}\) denote
passage-, sentence-, and triple-level evidence views, respectively.
\(\mathcal{I}\) is a cross-granularity evidence index that links sentence and
triple records to their source passages. Passage embeddings support dense
retrieval, after which the associated sentence and triple records are collected
through \(\mathcal{I}\) rather than retrieved independently.

The three views provide complementary support. Passages retain broad context for
entity disambiguation and cross-sentence reasoning. Sentences preserve local
constraints such as aliases, temporal cues, negation, and comparison. Triples
encode compact relational facts extracted by a large language model. MEGRAG
therefore accesses fine-grained evidence only within the retrieved passage set,
without relying on corpus-level knowledge-graph traversal.

\subsection{Path-structured Question-specific Evidence Graph Construction}

Given the Initial Query \(x\), MEGRAG collects evidence candidates, constructs a
reasoning node, and updates the path deterministically at each step. With a
maximum of \(B\) steps, the process continues until
\(q_{i+1}=\emptyset\) or \(i=B\).

\noindent\textbf{Candidate collection.}
At step \(i\), MEGRAG encodes \(q_i\) and ranks all passages by vector
similarity. The top \(N_P\) passages form \(C_i^P\). Rather than retrieving
fine-grained units independently over the full corpus, MEGRAG uses
\(\mathcal{I}\) to collect the sentences and triples associated with
\(C_i^P\).
Following the passage ranking, duplicate units are removed and the retained
sentence and triple candidates form \(C_i^S\) and \(C_i^T\), bounded by
\(N_S\) and \(N_T\). Specifically, MEGRAG traverses passages in retrieval-rank
order, preserves each passage's stored sentence and triple order, removes
duplicates, and retains the first \(N_S\) unique sentences and \(N_T\) unique
triples. Thus, \(C_i=\{C_i^P,C_i^S,C_i^T\}\) contains three aligned evidence
views derived from the same retrieved passages.

\noindent\textbf{Reasoning node construction.}
Before constructing \(v_i\), MEGRAG summarizes \(G^{(i-1)}\) as \(H_i\). For
each prior node and its outgoing edge, \(H_i\) records the current query,
selected evidence, intermediate answer, resolved path, missing information,
and goal-conditioned transition relation:
\(H_i=\mathrm{GraphSummary}(G^{(i-1)})\). The policy is
\begin{equation}
o_i=(Z_i,b_i,y_i,s_i,q_{i+1},h_i,m_i,r_i)
=\pi(x,q_i,H_i,C_i).
\end{equation}
The selected evidence is
\begin{equation}
Z_i=(T_i,S_i,P_i),
\end{equation}
where \(T_i\), \(S_i\), and \(P_i\) are the selected triple-, sentence-, and
passage-level evidence aligned through \(\mathcal{I}\). A node may contain one
or more granularities. MEGRAG evaluates evidence in the fine-to-coarse order
\(C_i^T\rightarrow C_i^T\cup C_i^S\rightarrow
C_i^T\cup C_i^S\cup C_i^P\). It first determines whether the triples are
sufficient for the current query. If not, it adds aligned sentences and then
passages until reaching the first sufficient granularity. The resulting
composition becomes \(Z_i\); unselected evidence is not carried into subsequent
reasoning. Based on \(Z_i\), the policy produces \(b_i\) and updates the resolved
path \(h_i\). If the accumulated findings do not yet resolve \(x\), it records
the missing information in \(m_i\), sets \(y_i=\emptyset\) and \(s_i=0\),
generates \(q_{i+1}\), and predicts \(r_i\). Otherwise, it sets
\(m_i=\mathrm{none}\), returns the answer as \(y_i\), sets \(s_i=1\), and
leaves \(q_{i+1}\) empty.

\noindent\textbf{Graph update and stopping.}
MEGRAG first appends \(v_i\). If \(s_i=0\), the next iteration constructs
\(v_{i+1}\) from \(q_{i+1}\) and adds the edge
\(e_i^{\mathrm{cont}}=(v_i,r_i,v_{i+1})\). Because each node produces at most
one next query, the graph is a directed path. The goal-conditioned relation
\(r_i\), defined from \(x\), \(q_i\), and \(q_{i+1}\), records how the next
retrieval step addresses what remains unresolved in the Initial Query. At a
valid stop, MEGRAG appends
\(e_i^{\mathrm{stop}}=(v_i,\mathrm{STOP})\) and returns
\begin{equation}
\hat{a}=y_i,\quad P_{\hat{a}}=\mathrm{Trace}(v_i,G^{(i)}).
\end{equation}
If the step budget is exhausted with \(s_B=0\), MEGRAG invokes a final
resolver using only the ordered selected evidence
\((Z_1,\ldots,Z_B)\), with \(b_B\) as a candidate clue. Its output is returned
as the prediction, while the trajectory remains marked as budget-exhausted.
At each step, the trace records the current query, selected evidence,
intermediate answer, resolved path, remaining information need, and, for a
continuing node, its outgoing transition relation.

\subsection{Graph Construction Distillation}

At each step, the graph construction policy \(\pi\) selects evidence, produces
an intermediate answer, checks whether the Initial Query has been resolved, and,
when necessary, generates the next query. MEGRAG distills this behavior into a
lightweight student policy.

For each training question, the teacher runs the full construction process and
produces a trajectory
\(\xi=\{(x,q_i,H_i,C_i,o_i)\}_{i=1}^{L_\xi}\). The student is trained to
reproduce the complete teacher decision \(o_i\) at each step:
\begin{equation}
\mathcal{L}_{\mathrm{GCD}}=-\sum_{\xi}\sum_{i=1}^{L_\xi}
\log p_{\theta}(o_i\mid x,q_i,H_i,C_i).
\end{equation}
Terminal decisions supervise \(s_i=1\), \(y_i\), and
\(m_i=\mathrm{none}\), together with \(q_{i+1}=\emptyset\). The deterministic
graph-update rule is unchanged.



\section{Experiments}

We evaluate answer accuracy (RQ1), component contributions (RQ2), adaptive
evidence granularity and retrieval depth (RQ3), and stopping efficiency (RQ4).
The supplement provides extended results, implementation details, an evidence
sufficiency audit, failure analysis, and a complete trajectory. Appendix~B.7
tests whether selected evidence alone can reproduce the answer.

\begin{figure*}[t]
    \centering
    \includegraphics[width=\textwidth]{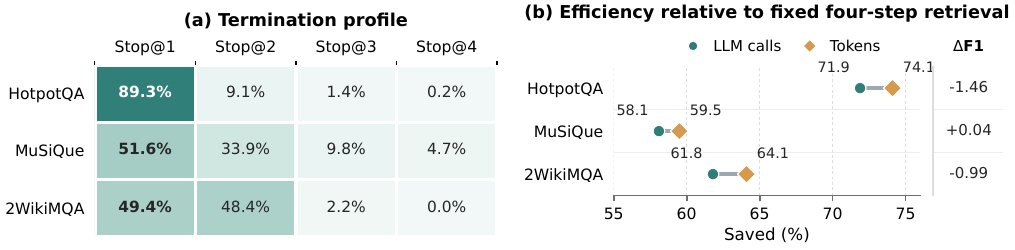}
    \caption{Termination behavior and measured efficiency relative to fixed four-step retrieval. Left: termination-step distribution. Right: saved LLM calls and tokens, together with the change in answer F1.}
    \label{fig:termination_efficiency}
\end{figure*}

\subsection{Experimental Setup}

\paragraph{Datasets and metrics.}
We evaluate on 2WikiMultiHopQA~\cite{ho2020constructing},
HotpotQA~\cite{yang2018hotpotqa}, and MuSiQue~\cite{trivedi2022musique}, reporting
token-level F1 and exact match (EM). Under both backbones, every method uses the
same seed-42 subset of 1,000 questions per benchmark, retrieval corpus,
preprocessing, and evaluation scripts. Questions are sampled without
replacement. Unless noted, analyses use these subsets with Qwen3-8B.

\paragraph{Baselines.}
Grouping methods by their primary design emphasis, we compare with \emph{Direct Model} and \emph{NativeRAG}~\cite{lewis2020retrieval}; the iterative or adaptive methods \emph{MetaRAG}~\cite{zhou2024metacognitive}, \emph{DualRAG}~\cite{cheng2025dualrag}, and \emph{CIRAG}~\cite{wei2026cirag}; the structure-aware methods \emph{HippoRAG~2}~\cite{gutierrez2025rag}, \emph{NeocorRAG}~\cite{peng2026neocorrag}, \emph{HGRAG}~\cite{wang2026cross}, \emph{LogicRAG}~\cite{chen2026you}, and \emph{QAFD-RAG}~\cite{zhou2026query}; and the multi-granular method \emph{MGranRAG}~\cite{hu2026iterative}.
Each baseline follows its published training regime: prescribed training stages
are reproduced, while training-free methods remain training-free. Therefore,
the Qwen3-8B results compare complete systems rather than architectures under
matched supervision. We separately report MEGRAG without SFT and prompt-only
MEGRAG with Qwen3-Max to isolate the effects of distillation and the inference
procedure.

\paragraph{Implementation and Training Details.}

\textbf{Backbone.} We use Qwen3-8B-Instruct and
qwen3-max-2026-01-23~\cite{yang2025qwen3}, denoted Qwen3-8B and Qwen3-Max.
Within each setting, all methods share the reasoning and answer backbone;
Qwen3-Max also supplies offline OpenIE and distillation trajectories.

\textbf{Retrieval Setup.} We use nvidia/NV-Embed-v2~\cite{lee2025nv} as the
default retriever and BGE-small-en-v1.5~\cite{xiao2024c} for the retriever
robustness study. Both encode queries and passages. Sentence and triple
candidates are obtained by aligned lookup from the retrieved passages, not by
independent embedding retrieval. Iterative methods retrieve 10 passages per
step for at most four steps. For MEGRAG, \(N_P=10\) and the aligned candidate
budgets are \(N_S=N_T=30\).

\textbf{Distillation.} Qwen3-Max generates trajectories for 3,000 questions
from the official training splits, disjoint from evaluation, and Qwen3-8B is
fine-tuned with LoRA~\cite{hu2022lora} on one H800 GPU. Qwen3-8B results use the
distilled system, whereas Qwen3-Max applies the same inference procedure
prompt-only. The w/o-SFT ablation measures the effect of distillation.
Appendix~A gives trajectory filtering and optimization details.

\subsection{Main Results on Multi-hop QA}

Table~\ref{tab:main_multihop} answers RQ1. MEGRAG has the highest observed F1
and EM in all 12 comparisons. With Qwen3-8B, its F1/EM margins over the
strongest per-dataset baseline are 5.84/7.50, 2.90/2.70, and 8.34/9.10 on
2WikiMultiHopQA, HotpotQA, and MuSiQue. These end-to-end comparisons use each
system's native training regime and do not attribute the entire margin to
architecture alone. The largest gains occur on MuSiQue, whose questions
generally require longer compositional chains.

With prompt-only Qwen3-Max, the respective margins remain 1.92/1.10,
3.34/1.20, and 5.21/6.30, showing that the inference procedure remains
effective without trajectory distillation. Table~\ref{tab:ablation} measures
the additional effect of distillation within MEGRAG. Paired-bootstrap intervals
in Appendix~B.1 confirm both metrics for every Qwen3-8B comparison. With
Qwen3-Max, all F1 gains and both MuSiQue gains are significant, whereas EM on
2WikiMultiHopQA and HotpotQA is not. We therefore report the observed rankings
without claiming uniform significance.

\subsection{Goal-conditioned Transition Analysis}

To isolate the transition relation from trajectory distillation, we conduct a
focused prompt-only Qwen3-Max ablation on MuSiQue.

\begin{center}
\begin{minipage}{\columnwidth}
\centering
\small
\renewcommand{\arraystretch}{1.08}
\begin{tabular*}{\columnwidth}{@{\extracolsep{\fill}}lrrr@{}}
\toprule
\textbf{Variant} & \textbf{F1} & \textbf{EM} & \(\Delta\)\textbf{F1} \\
\midrule
\textbf{Full MEGRAG} & \textbf{62.01} & \textbf{51.80} & -- \\
w/o Transition Relation & 60.40 & 50.20 & $-1.61$ \\
Shuffled Transition Relation & 59.10 & 48.60 & $-2.91$ \\
\bottomrule
\end{tabular*}
\captionof{table}{Effect of goal-conditioned transition relations on MuSiQue with
prompt-only Qwen3-Max. All variants use the same retriever, candidate budgets,
reasoning-state fields, stopping policy, and inference budget; only the
provided transition relation is modified.}
\label{tab:transition_relation}
\end{minipage}
\end{center}

\emph{w/o Transition Relation} removes \(r_i\) but retains all other state
fields; \emph{Shuffled Transition Relation} replaces it with an unrelated
relation. The F1/EM losses are 1.61/1.60 and 2.91/3.20, respectively. The larger
loss after shuffling suggests that later decisions depend on the content of the
relation, not merely its presence. Because all variants are prompt-only, the
differences cannot be attributed to trajectory fine-tuning. We focus on
MuSiQue because its longer compositions directly test whether transition
information helps track what remains unresolved across hops.

\begin{table*}[t]
\centering
\small
\renewcommand{\arraystretch}{1.10}
\begin{tabular*}{\textwidth}{@{\extracolsep{\fill}}lcccccc@{}}
\toprule
\textbf{Method}
& \multicolumn{2}{c}{2WikiMQA}
& \multicolumn{2}{c}{HotpotQA}
& \multicolumn{2}{c}{MuSiQue} \\
\cmidrule(lr){2-3}\cmidrule(lr){4-5}\cmidrule(lr){6-7}
& \textbf{F1} & \textbf{EM}
& \textbf{F1} & \textbf{EM}
& \textbf{F1} & \textbf{EM} \\
\midrule
\textbf{MEGRAG}
& \textbf{76.34} & \textbf{68.70} & \textbf{76.95} & \textbf{61.90} & \textbf{55.15} & \textbf{44.90} \\
\midrule
MEGRAG w/o SFT
& 57.93 & 50.60 & 70.62 & 55.70 & 43.58 & 31.80 \\
Triples only
& 61.20 & 55.40 & 68.33 & 53.90 & 43.35 & 33.00 \\
Sentences only
& 73.51 & 64.70 & 72.38 & 56.04 & 48.92 & \underline{39.05} \\
Passages only
& \underline{74.89} & \underline{67.00} & 73.46 & 57.32 & 43.96 & 34.40 \\
w/o Structured History
& 74.85 & 66.80 & \underline{74.12} & \underline{58.66} & \underline{49.34} & 38.60 \\
\bottomrule
\end{tabular*}
\caption{Ablation study on multi-hop QA benchmarks with Qwen3-8B-Instruct. The best result in each column is shown in bold, and the strongest ablated variant is underlined.}
\label{tab:ablation}
\end{table*}

\subsection{Ablation Study}

Table~\ref{tab:ablation} addresses RQ2. Removing SFT reduces F1 by 18.41,
6.33, and 11.57 points, showing its importance for transferring structured
decisions to Qwen3-8B. This does not imply that MEGRAG itself requires
fine-tuning: prompt-only Qwen3-Max retains the advantage in
Table~\ref{tab:main_multihop}.

No fixed granularity dominates: passages are strongest on 2WikiMultiHopQA and
HotpotQA, whereas sentences are strongest on MuSiQue. MEGRAG exceeds these
per-dataset best variants by 1.45, 3.49, and 6.23 F1, supporting adaptive
composition: triples provide compact evidence, while aligned sentences and
passages restore context and qualifiers when needed. The gain therefore
reflects adaptive granularity selection rather than reliance on one evidence
view.

\emph{w/o Structured History} keeps the candidates, policy, budget, and
stopping rule, but replaces prior queries, intermediate answers, resolved
paths, missing information, and transitions with accumulated evidence alone.
Its F1 drops by 1.49, 2.83, and 5.81, showing that the combined reasoning
history is especially useful on MuSiQue. Because the variant removes all
history fields together, it measures their joint contribution rather than that
of any single field.

\subsection{Adaptive Behavior Across Question Structures}

\begin{figure}[t]
    \centering
    \includegraphics[width=\columnwidth]{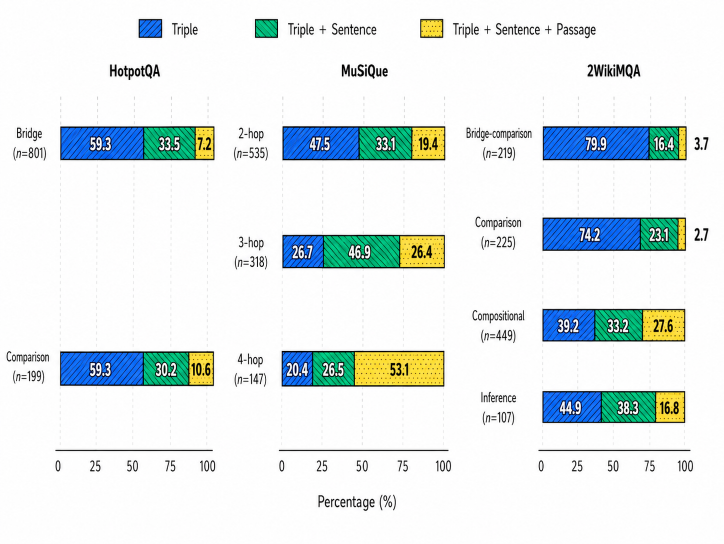}
    \caption{Evidence composition across question structures. Each bar reports the fraction of questions using triples only, triples with sentences, or all three evidence granularities.}
    \label{fig:evidence_composition}
\end{figure}

\begin{figure}[t]
    \centering
    \includegraphics[width=\columnwidth]{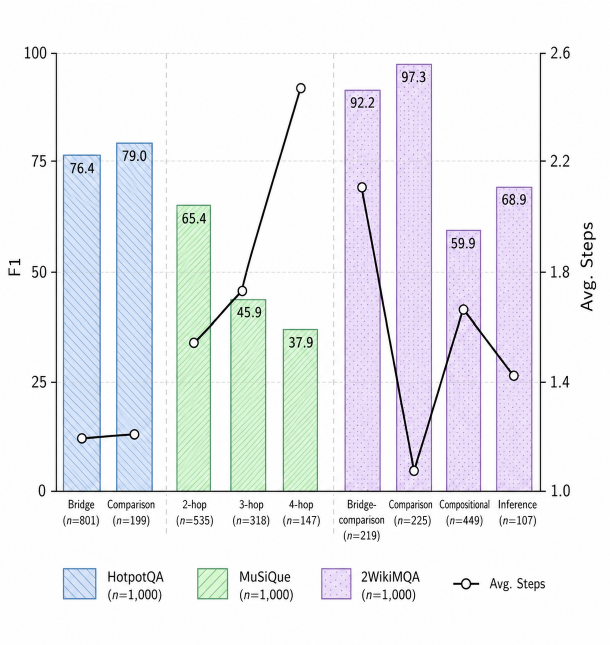}
    \caption{Answer F1 and average retrieval steps across question structures. Bars report F1 and the black line denotes average retrieval depth.}
    \label{fig:f1_steps}
\end{figure}

Figures~\ref{fig:evidence_composition} and~\ref{fig:f1_steps} address RQ3 by comparing evidence composition and retrieval depth across question structures.

On MuSiQue, use of all three granularities rises from 19.44\% at two hops to
53.06\% at four, and average depth from 1.447 to 2.497; F1 nevertheless falls
from 65.4 to 37.9, confirming that longer chains remain harder. On
2WikiMultiHopQA, MEGRAG takes more retrieval steps for bridge-comparison
questions, which remain triple-dominant, while compositional questions use
passages more often.
HotpotQA structures are largely triple-dominant and shallow. Evidence
granularity and retrieval depth therefore vary independently with the
information need. Harder questions tend to trigger richer evidence and more
retrieval steps, yet longer chains remain challenging.
The cross-dataset contrast also indicates that hop count alone does not
determine granularity; the form of the missing evidence matters.

\subsection{Robustness and Sensitivity}

Appendix~B.2--B.5 tests retriever, backbone, and budget robustness. Replacing
NV-Embed-v2 with BGE-small-en-v1.5 preserves MEGRAG's highest observed F1 and
EM in all six comparisons. Across Llama, DeepSeek, GPT, and Gemini, it also
ranks first in all 24 comparisons, with the largest margins on MuSiQue. These
results extend the two main Qwen settings without changing the evaluation
protocol.

On MuSiQue, increasing the step budget from one to four raises F1 from 48.50
to 55.15 and reduces budget exhaustion from 48.4\% to 2.8\%, while average
depth grows only from 1.000 to 1.676. Gains diminish after three steps. The
default candidate budget improves F1 by just 1.03--1.24 over the smallest
tested pair, and larger pools add no gain, indicating that performance is not
sharply tuned to the candidate budget. The fourth step therefore accommodates
the remaining difficult chains without forcing four steps on every question.

\subsection{Answer-aware Stopping and Efficiency}

\begin{figure}[t]
    \centering
    \includegraphics[width=\columnwidth]{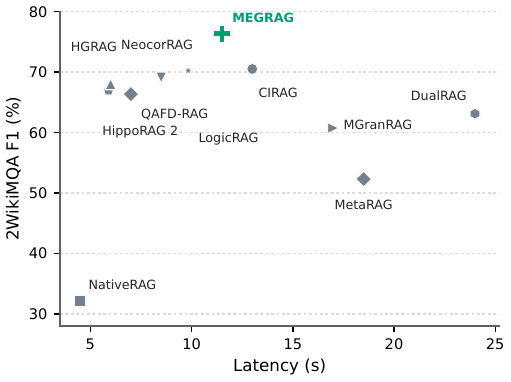}
    \caption{End-to-end latency versus F1 on 2WikiMultiHopQA with
    Qwen3-8B-Instruct.}
    \label{fig:latency_main}
\end{figure}

Figure~\ref{fig:termination_efficiency} compares answer-aware stopping with
fixed four-step retrieval; $\Delta$F1 is their F1 difference. Since each step
answers the current query, the controller can stop when the accumulated state
resolves the Initial Query rather than retrieve to a fixed depth. HotpotQA stops
after one step for 89.30\% of questions, saving 71.88\% of LLM calls and
74.10\% of tokens at $-1.46$ F1. MuSiQue continues more often, saving
58.10\%/59.50\% of calls/tokens with $\Delta$F1 \(=+0.04\).
2WikiMultiHopQA saves 61.80\%/64.10\% at a 0.99-point cost. Thus, stopping
adapts computation to unresolved information rather than maximizing early
termination. Stopping behavior also varies across datasets, indicating that the
controller responds to different information needs.
Figure~\ref{fig:latency_main} reports wall-clock latency and F1 on
2WikiMultiHopQA. MEGRAG achieves the highest F1 (76.34) with an end-to-end
latency of 11.5 seconds. Although slower than several graph-based baselines, it
is faster and more accurate than CIRAG, MGranRAG, MetaRAG, and DualRAG, yielding
a favorable accuracy--latency trade-off. Appendix~B.4 gives complete
termination and efficiency statistics.

\subsection{Transfer to Single-hop QA}

On single-hop NQ~\cite{kwiatkowski2019natural} and
WebQ~\cite{berant2013semantic}, MEGRAG leads the strongest baseline by
0.39/1.90 and 1.50/2.15 F1/EM, respectively (Appendix~B.6). These English
factoid datasets test whether iterative reasoning degrades performance when no
long retrieval path is required; MEGRAG remains competitive in this setting.

\subsection{Case Study}

Appendix~C traces a three-hop MuSiQue example. MEGRAG first resolves an
intermediate relation with compact evidence, then expands to aligned context
to resolve the remaining reference and answer the Initial Query. The trace
shows how selected granularity, the intermediate answer, and the next-query
decision interact along one path.

\section{Conclusion}

MEGRAG combines indexed multi-granular evidence with question-specific
iterative reasoning. At each step, it selects evidence at the first granularity
judged sufficient, produces an intermediate answer, and uses the remaining
information need to decide whether to continue retrieval. Experiments on three
multi-hop QA benchmarks and two backbones show consistent gains over diverse
RAG baselines. Further analyses support the contributions of adaptive evidence
selection, reasoning history, transition information, policy distillation, and
answer-aware stopping.

\section{Limitations}

MEGRAG cannot recover evidence outside its passage-retrieval scope, and
automatically extracted triples may contain errors or omit qualifiers. Its
single path cannot backtrack after a wrong intermediate state. The lightweight
Qwen3-8B setting also requires task-specific teacher trajectories, whereas
Qwen3-Max does not. Multi-hop evaluation uses benchmark-specific candidate
corpora, and reported latency excludes one-time OpenIE, embedding, and indexing
costs. Because each node has at most one successor, MEGRAG should be understood
as a path-based reasoning framework rather than a general graph-search or
message-passing method. Paired bootstrap captures uncertainty within each fixed
subset, not variation across independently sampled subsets. Evaluation beyond English
factoid QA and the fixed triple-first selection order remains future work.

\bibliography{aaai2027}
\end{document}